\documentclass{isprs} %
\usepackage{subfigure}
\usepackage{setspace}
\usepackage{geometry} %
\usepackage{epstopdf}
\usepackage[labelsep=period]{caption}  %
\usepackage[british]{babel} 
\usepackage[hang]{footmisc}

\usepackage{amsmath,epsfig}
\usepackage{booktabs}
\usepackage{multicol,multirow}
\usepackage{xcolor}
\usepackage{comment}
\usepackage{url}
\usepackage{bbm}
\usepackage{makecell}
\usepackage{natbib}

\graphicspath{{Images/}}

\newcommand{\alex}[1]{{\color{black} #1}}

\DeclareMathOperator*{\argmax}{argmax}

\begin{document}

\title{CroCo: Cross-Modal Contrastive learning for localization \\ of Earth Observation data}

\date{}

\author{
    Wei-Hsin Tseng$^{1}\qquad$
    Hoàng-Ân Lê$^{1,}$\thanks{Corresponding author. E-mail:  \protect\url{hoang-an.le@irisa.fr}}$\qquad$
    Alexandre Boulch$^{2}\qquad$
    Sébastien Lefèvre$^{1}\qquad$
    Dirk Tiede$^{3}$
}
\address{
    $^{1}$ IRISA, Université Bretagne Sud, France \\ %
    $^{2}$ Valeo.ai, France \\ %
    $^{3}$ Department of Geoinformatics - Z\_GIS, University of Salzburg, Austria
}

\icwg{II/III}   %

\abstract{
    It is of interest to localize a
ground-based LiDAR point cloud on remote sensing imagery.
In this work, we
tackle a subtask of this problem, i.e. to map a digital elevation model (DEM)
rasterized from aerial LiDAR point cloud on the aerial imagery. We proposed a
contrastive learning-based method that trains on DEM and high-resolution
optical imagery and
experiment the framework on different data sampling strategies and hyperparameters.
In the best scenario, the Top-1 score of 0.71 and Top-5 score of 0.81
are obtained. The proposed method is promising for feature learning from RGB and DEM for localization and is potentially applicable to other data sources too.  
Source code is released at~\url{https://github.com/wtseng530/AVLocalization}.

}

\keywords{Data fusion, Contrastive Learning, DEM, Aerial Imagery, Localization.}

\maketitle
\sloppy

\section{Introduction} \label{sec:intro}

The loss of geo-localization information can be critical for various applications from discarded data (processing time/money loss) to autonomous vehicles stopping due to planning failures.
One possible solution to this problem is to rely on a fallback solution such as retrieving the location of the sensor by identifying the acquired data to a known, large-scale, geo-localized image map.
In this study, we consider the retrieval of 3D-patch locations in large RGB scenes.
The choice of these two modalities is driven by practical applications.
\paragraph{Target RGB data.} First, we chose the large scale data to be RGB, because it is among the most abundant available data.
For example, services such as Google Maps\footnote{\url{https://maps.google.com/}} or Bing Maps\footnote{\url{https://www.bing.com/maps}}  are providing high quality RGB images of the entire Earth.
These are geo-referenced and can directly be linked with mapping data such as OpenStreetMaps\footnote{\url{https://www.openstreetmap.org}}. The reason for not directly trying to localize in abstract maps is that this type of data, while semantically complete, lacks real geometric information, helpful for geo-localization.
\paragraph{Lidar data.}
For the source data, we chose a 3D modality, such as lidar data. This modality covers mainly two possible acquisition sources, namely aerial lidar (a) and autonomous driving lidar (b). As stated before, the loss of GPS information during the acquisition can lead to post-processing difficulties, long manual data registration, thus, human workforce and money; as well as large range of consequences, from car stopping and traffic jam inconvenience to car accident.
For the particular case of autonomous driving, 3D data has advantages over ground RGB to contain the geometry, thus it is easier to change the point of view (move to bird-eye view, same as target data). Moreover, it is less subject to weather/time-of-the-day changes, e.g., day vs night.

One arising difficulty in using 3D data and RGB images, is that identifying them requires extracting the common information they contain.
Common feature extractors, such as SIFT~\citep{Lowe2004}, HOG~\citep{Dalal2005}, pretrained ImageNet models~\citep{imagenet}, etc., would fail, because there is no guaranty that different inputs will produce the same descriptor.

To tackle this problem, we propose jointly training two neural networks, one for RGB processing, the other for 3D processing:
the two networks learn to produce similar responses to inputs of the same locations, and different for different locations.
To that end, the contrastive loss introduced in SimCLR~\citep{Chen2020} is used to force a high feature correlation for positive data pairs,~\emph{i.e.} from the same locations, and low correlation for negative pairs.
The resulting approach directly learns from co-registered RGB-lidar data without any additional annotation.

Additionally, to assess the validity of the proposed method, an experimental setup is designed to relocalize elevation patches in RGB data using
the IEEE GRSS Data Fusion Contest 2018 dataset~\citep{Xu2019}.
The dataset contains co-located aerial images and aerial lidar scans, rasterized into elevation models and split into geographically separated training and testing data.
Although elevation model is built from aerial lidar (a), it is also closely related to autonomous driving lidar (b) in the sense that elevation maps are commonly produced for automotive trajectory planning and detection.

Our contributions are thus as follow:
\begin{enumerate}
    \item We propose a contrastive approach inspired from self-supervision models for cross-modality patch registration in large images; 
    \item We define, based on the DFC 2018 dataset, a 3D patch retrieval in RGB aerial images experimental setup;
    \item We quantitatively and qualitatively assess that such a learning scheme can learn correlated features for cross-modal localization.
\end{enumerate}

The paper is structured as following: Section~\ref{sec:related_work} is dedicated to previous and related work, Section~\ref{sec:method} describes the network architecture, Section~\ref{sec:dataset} presents the dataset and Section~\ref{sec:experiments} exposes the experimental results.

\section{Related Work} \label{sec:related_work}

Visual place recognition retrieves the coarse location of a target in a known
scene by matching query input to a pre-built map.
It is commonly approached in a
two-stage manner -- mapping and localizing~\citep{Bansal2011,Lin2015,Noda2011,Sunderhauf2015,Workman2015}. In the
mapping stage, images or point clouds are represented by a collection of local
features. These local features are then aggregated into global features, which
join the corresponding camera pose or geo-referencing and forms a prior map. In
the localization stage, the query data is compared with the pre-built map, and
the most relevant part of the scene is retrieved as the mobile agent’s location~\citep{Chen2020}.

Aerial and satellite imagery have become a popular source for coarse level
localization and place recognition~\citep{Bansal2011,Chu2015,Kim2017,Workman2015}. It resolves the challenges of
viewpoint variations and appearance variations that occur in ground-based
imagery~\citep{Kim2017}.

\citet{Lin2015} are the first to propose a deep learning method for
cross-view geo-localization using Siamese CNN. Their Where-CNN model shows a significant improvement over the handcrafted features.
\citet{Workman2015wide} match the query ground images against an aerial image database,
represented by features extracted from a pre-trained model~\citep{Workman2015}. Another ground to overhead image matching explores different deep learning approaches for matching and retrieval tasks~\citep{Vo2016}.
The authors propose a new loss function and incorporate rotation invariance and
orientation regression during training. In~\citep{Cai2019}, a hard
exemplars mining strategy for cross-view matching outperforms the soft-margin
triplet loss. Two CNNs are used to transform ground-level and aerial images into
a common feature space. And the query ground image is localized by matching the
closest geo-referenced aerial image.

An important hypothesis on which contrastive learning is
built is the fact that related samples have closer embedding space. It learns
feature space (or encoder) that joins related points and apart unrelated ones.
Definition of related/positive and unrelated/negative is determined based on the
context and objective. Positive and negative pairs can be nearby patches versus
distant patches~\citep{Oord2019} or patches from the same image versus from
different images. Contrastive loss is computed by contrasting latent
representations of positive and negative samples. Oord et al. proposed the Noise
Contrastive Estimation (NCE) which considers a random subset of negative samples
for loss computation to reduce expensive computation from many negative samples~\citep{Oord2019}.

SimCLR is a modified self-supervised learning algorithm for visual
representation~\citep{Chen2020}. It learns representations by maximizing the
agreement between the same data example with different augmentation operations
(positive pair), via a contrastive loss in the latent space.  

In our work, we adapt the SimCLR architecture for a different purpose.
The original SimCLR uses augmented generic images (ImageNet) to pre-train the feature encoder
for downstream tasks. For us, we train on aerial imagery and elevation data
without data augmentation operations. After training, the feature encoder is
used directly to extract feature descriptors for localization.

\section{Method} \label{sec:method}

In this section, we detail the proposed approach for learning patch relocalization.

\subsection{Localization Workflow}

Our approach, Cross-Modal Contrastive learning for localization of Earth Observation data (CroCo), uses a two-stage visual place framework.

\paragraph*{Mapping step.}
We first create a feature map of the target, large-scale, RGB data.
This map will be then used as a prior map for location queries.
The features are computed using a convolutional neural network which takes as input RGB patches generated with a sliding window over the original RGB image.

Let $\phi_\text{RGB}$ be the transfer function of the RGB network and $\mathcal{I}_\text{RGB}$ be the set of patches created with the sliding window, then:
\begin{equation}
    \mathcal{Z}_\text{RGB} = \left\{ z_\text{RGB}^i = \phi_\text{RGB}\left(x_\text{RGB}^i\right) \left| \;  x_\text{RGB}^i \in \mathcal{I}_\text{RGB} \right.\right\}
\end{equation}
is the set of feature vectors that will be used for localization.

\paragraph*{Localization step.}
The second stage aims at localizing the source 3D image patch in the target RGB image.
To do so, we use a second CNN, to produce a descriptor of the 3D data.
This descriptor is a feature vector similar to those generated at the first stage.

Let $\phi_\text{3D}$ be the transfer function of this second CNN.
Let $x_\text{3D}$ be the 3D input data, then:
\begin{equation}
    z_\text{3D} = \phi_\text{3D}\left(x_\text{3D}\right)
\end{equation}
is the descriptor of the input data.

Finally, the localization step is the search for the maximally correlated feature in $\mathcal{Z}_\text{RGB}$,
\begin{equation}
    i^* = \argmax_{i\;\text{s.t.}\;z_\text{RGB}^i \in \mathcal{Z}_\text{RGB}} \frac{ {z_\text{RGB}^{i}}^\top z_\text{3D}}{\left\|z_\text{RGB}^i\right\|_2 \left\|z_\text{3D}\right\|_2}
\end{equation}

\subsection{Cross-modal contrastive learning}

\begin{figure} \centering
\includegraphics[width=\linewidth]{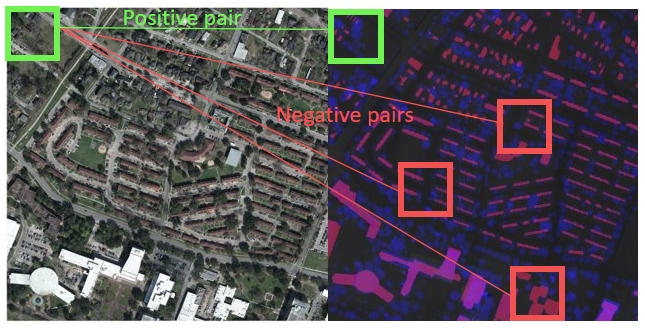} \caption{Positive  and negative pairs on RGB \& DEM imagery.}
\label{fig:positive_negative} \end{figure}

For the previous framework to be efficient, $\phi_\text{3D}$ and $\phi_\text{RGB}$ must be trained jointly to produce meaningful features,~\emph{i.e.}, features that are similar when the inputs come from the same location, and different in the other case.
The goal is then to produce highly correlated features for input positive input pairs (same location) and contrasted features for negative pairs (different locations), as shown in Fig.~\ref{fig:positive_negative}.

Contrastive learning is widely used in self-supervised learning.
In this work we use the NT-Xent loss (the normalized temperature scaled cross-entropy loss), introduced in SimCLR~\citep{Chen2020}:
\begin{equation}
    l_{i,j} = -\log\dfrac{\exp\left(\text{sim}\left(z_i,z_j\right)/\tau\right)}{\sum_{k=1}^{2N}{\mathbbm{1}_{[k\neq i]}\exp\left(\text{sim}\left(z_i,z_k\right)/\tau\right)}}
\end{equation}
where $\left(z_i, z_j\right)$ is a positive pair of examples, $\tau$ denotes a temperature parameter and
$\text{sim}\left(z_i,z_j\right)$ is the dot product between two normalized representations $z_i$ and $z_j$:
\begin{equation}
    \text{sim}\left(u,v\right) = \dfrac{u^\top v}{\left\|u\right\|\left\|v\right\|}
\end{equation}

\subsection{Network architecture}

\begin{figure}[t] \centering
\includegraphics[width=.7\linewidth]{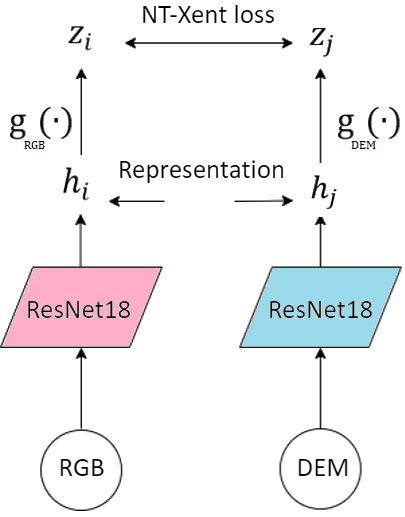} \caption{Proposed
contrastive learning framework for VPR.} \label{fig:proposed_vpr} \end{figure}

The proposed architecture is presented in Fig.~\ref{fig:proposed_vpr}.
Each branch (RGB and elevation) is composed of a convolutional backbone and a projection head.
It is similar in design to the one presented in SimCLR.

However, as our goal is to learn a cross-modal contrastive loss, each branch processes different type of data.
Therefore, we do not follow a Siamese approach, and the two branches are independent (not sharing weights). 
Moreover, as opposed to self-supervised training, which aims at training the convolutional backbone and discards the projection head for downstream tasks, we use the full network at inference time (backbone and projection head).

\emph{Implementation details.}
In our experiments, we use a ResNet18 convolutional backbone and the 1-layer MLP as a projection head to map features to a
128-dimensional latent space.
Please note that any convolutional backbone could be used.
We chose this architecture for its relatively low memory footprint.

\subsection{Training CroCo}

At training time, we adapt the training procedure by \citet{Chen2020} to cross-modal data.

In each step, $N$ random positive data pairs are sampled.
From these positive pairs, we build the
$N(N-1)$
possible negative pairs.
Each data points have one positive pair and
$(N-1)$
negative pairs. 

In practice, for each data input, we compute its corresponding feature vector using $\phi_\text{RGB}$ or $\phi_\text{3D}$ and the NT-Xent loss is applied for all possible data pairs (positive and negative) in the data batch.

\section{Dataset}
\label{sec:dataset}

To the best of our knowledge, the proposed task, 3D data relocalization in large-scale images, has not been tackled before. Therefore, no dedicated pre-existing dataset could be found.
In order to qualitatively and quantitatively assess the performances of CroCo, we repurpose the dataset GRSS\_DFC\_2018\footnote{2018 IEEE GRSS Data Fusion Contest~\citep{Xu2019}.
Online: \url{http://www.grss-ieee.org/community/technical-committees/data-fusion}}. 

\begin{figure}
    \centering
    \includegraphics[width=\linewidth]{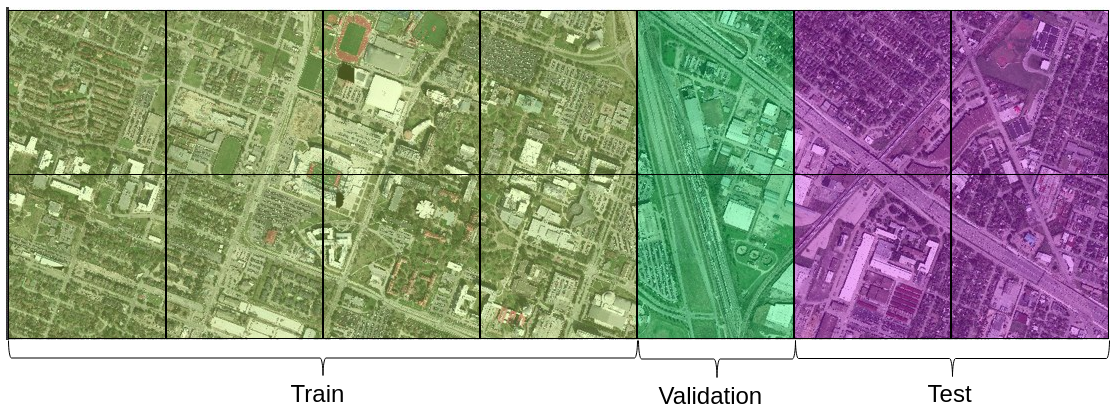}
    \caption{Train (8 tiles) / Validation (2 tiles) / Test (4 tiles) split on DFC
        dataset.}
    \label{fig:DFC_splits}
\end{figure}

\paragraph*{2018 Data Fusion Contest (DFC) dataset.}
The DFC dataset~\citep{Xu2019} dataset by the IEEE Geoscience and Remote Sensing Society consists of multi-sensor data covering 4~km$^2$
of the University of Houston campus, which is divided into 14 image tiles, each accounting for $601\times596$ m$^2$.
The predefined tiles are split into a training set (8 tiles), validation set (2 tiles), and testing set (4 tiles) as shown in Fig.~\ref{fig:DFC_splits}.

\paragraph*{Modality creation.}
The original dataset is designed for multi-sensor fusion and
land cover classification task. We make use of the paired RGB imagery
and DSM data to train a cross-modal localization model.

\begin{figure}[t] \centering
    \includegraphics[width=\linewidth]{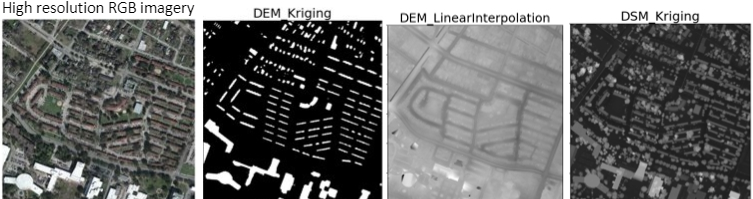}
    \includegraphics[width=\linewidth]{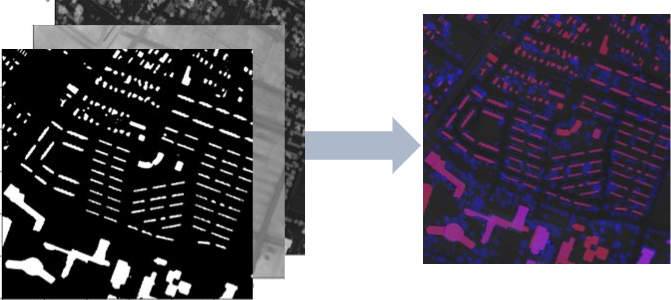} \caption{DEM
    dataset created from DEM kriging, DEM triangulation, and DSM.}
    \label{fig:DEM_dataset} \end{figure}

In the original dataset, four digital surface model products derived from LiDAR data are provided, namely (1) first surface model (DSM) generated from first returns, (2) bare-earth digital elevation model (DEM) from returns classified as ground, (3) bare-earth DEM with void filling for man-made structures, and (4) a hybrid ground and building DEM.
We opt for the (1), (3), and (4) and concatenate them as in Fig~\ref{fig:DEM_dataset}.

\alex{
As RGB imagery is provided in very high resolution (5cm GSD), we first need to align optical and elevation data.
}

\alex{
\paragraph*{Objective tasks.}
In order to correctly assess the performance of localization in a realistic scenario, we consider that localization is good if we retrieve the location with a given maximal error.

In this approach, we tackle this objective as a classification problem.
The RGB images are split in a collection of patches using a sliding window of 2m (resp. 4m).
The task is then given a 3D patch, find its corresponding patch in the image collection.

\paragraph{Metrics.}
Following the SimCLR by \citet{Chen2020}, we report Top-1 and Top-5 score. Top-1 score is the proportion of tests when the most confident retrievals,~\emph{i.e.}~those predicted with highest probability, are the target patches (values between $[0, 1]$, the higher the better). Similarly, Top-5 score is the proportion of tests when the \emph{five} most confident results contain the target, thus is more relaxed than Top-1.
}

\section{Experiments}
\label{sec:experiments}

\alex{
In this section, we evaluate how a contrastive learning framework, CroCo, can generate consistent features from RGB and DEM image pairs.
For the two variants of the dataset, 2m and 4m localization, we study the factors that may impact the performance of CroCo: ground sample distance (Sec~\ref{sec:gsd}), patch dimension (sec~\ref{sec:pd}) and batch size (sec~\ref{sec:bs}).
The first two factors determine the informative detail of the input imagery,
while the last is well-known for its impact in SimCLR~\citep{Chen2020}.
All these ablations are done on the validation set, as defined in the dataset section.
Finally, given the set of parameters, we evaluate our model on the test set.
}

\subsection{Ground sample distance}
\label{sec:gsd}

\alex{
As RGB imagery is provided at very high ground sample distance (5cm GSD), to align optical
and elevation data, we downsample the RGB images to lower resolution. Two scenarios
are tested, (1) RGBs are down-sampled to 50cm GSD while DEMs are kept at 50cm GSD,
and (2) RGBs and DEMs are down-sampled to 100cm GSD. The results with patch
dimension of 1024 m$^2$, sample stride of 4, and batch size of 256 are shown in
Tab.~\ref{tab:res50_100_2}.
It could be seen that the higher GSD results in higher accuracy as it provides
more details of the scene.
}

\begin{table}[t]
    \centering
    \begin{tabular}{@{}ccccc@{}}
    \toprule
    \multirow{2}{*}{\makecell[c]{Resolution\\(cm)}} & \multicolumn{2}{c}{2m loc.} & \multicolumn{2}{c}{4m loc.}\\
    \cmidrule(r){2-3}\cmidrule(l){4-5}
    & Top-1 & Top-5 & Top-1 & Top-5 \\
    \midrule
    50 & 0.43  & 0.57 & 0.40 & 0.55 \\
    100 & 0.37  & 0.51 & 0.40 & 0.52  \\
    \bottomrule
    \end{tabular}
    \caption{\alex{Ground sample distance study for 2m localization tasks with patch dimension of 1024 m$^2$ and batch size of 256.}}
    \label{tab:res50_100_2}
\end{table}

\subsection{Patch dimension}
\label{sec:pd}

\alex{
Patch dimension determines the spatial extension of each input tile.
We consider the context of autonomous driving, in which the LIDAR sensor can receive signals up to a hundred meters. As such, 2 patch sizes are tried: 256 m$^2$, (\emph{i.e.} $16\times 16$m$^2$) and 1024 m$^2$ (\emph{i.e.} $32\times 32$ m$^2$). The results with GSD of 50cm, for the 2m localization task, and batch size of 256
are shown in Tab.~\ref{tab:patchdim256_1024_2}. It could be seen that, larger patch dimension leads to better performance, as small patch size coverage leads to more patches with homogeneous appearances, hence less discriminative.
This suggests that identifying an appropriate patch dimension is critical for this approach. It should be selected according to the target task.     
}

\begin{table}[t]
    \centering
    \begin{tabular}{@{}ccc@{}}
    \toprule
    Patch dim (m$^2$) & Top-1 & Top-5 \\
    \midrule
    256 &  0.19 &  0.33 \\
    1024 & 0.43 & 0.57\\
    \bottomrule
    \end{tabular}
    \caption{\alex{ Patch dimension study for 2m localization tasks with resolution of 50cm and batch size of 256.}}
    \label{tab:patchdim256_1024_2}
\end{table}

\subsection{Batch size}
\label{sec:bs}

\alex{
In contrast to supervised learning, contrastive learning benefits
from the larger batch sizes. As already mentioned, each sample
instance in a batch of $N$ samples generates one positive pair and $(N-1)$ negative
pairs. More negative pairs available during training facilitate the convergence.
We vary the batch size $N$ from 128 to 512. We adopt the LARS optimizer for all
batch sizes used in SimCLR, it alleviates the unstableness yield from SDG/Momentum with linear learning rate scaling. The results are provided in  Tab.~\ref{tab:batchsize_res50_patch1024_2}.

In the SimCLR framework, the large batch size has
been strengthened to be an important factor for contrastive learning to perform
well. This is because the more samples there are in one batch, the more
negative pairs we will obtain for loss computation. It takes fewer epochs and
steps to get the same accuracy with the higher batch size. In the original
paper, the training batch size is set from 256 to 8192. With a limited memory
resource, we have to constrain our batch size to be less than 512. In general,
we can observe the tendency of rising accuracy with the batch size increment.
For patch size 512 m$^2$, best Top-1 and Top-5 scores are reached mostly in the highest batch size scenario.  
}
 
\begin{table}[t]
    \centering
    \setlength{\tabcolsep}{3pt}
    \begin{tabular}{@{}ccccc@{}}
    \toprule
    Batch size & Top-1 & Top-5 \\
    \midrule
    128 & 0.34  & 0.50\\
    256 & 0.40  & 0.55\\
    512 & 0.48  & 0.62\\
    \bottomrule
    \end{tabular}
    \caption{Batch size study for 4m localization task with resolution of 50cm and patch dimension of 1024 m$^2$. Larger batch size results in better performance.}
    \label{tab:batchsize_res50_patch1024_2}
\end{table}

\subsection{Evaluation on the test set}

\alex{
In Tab.~\ref{tab:results_2}, we present the results obtained on the test set.
From the ablation, we chose to use the 50cm resolution data, with a patch size of 64$\times$64 (1024m$^2$) and a batch size of 512 (the highest we can afford with our limited computational resources).

We obtain similar scores for both tasks with the same training procedure.
Intuitively, the 4m localization task with higher coverage would seem easier.
The main reason for not performing better in this setting resides in the way we generated the training data, following the same procedure as for the test data.
Thus, using a 2m sliding window produces more training patches than for 4m.
It appears that the size of the training set is also a parameter to be taken into account. We discuss it further in the discussion (Sec.~\ref{sec:discuss}).
}

\begin{table}[t]
    \centering
    \begin{tabular}{@{}ccc@{}}
    \toprule
    Localization tasks & Top-1 & Top-5 \\
    \midrule
    2m & 0.64 & 0.80 \\
    4m & 0.63 & 0.78 \\
    \bottomrule
    \end{tabular}
    \caption{Evaluation on the test set with ground sample distance of 50cm, patch dimension of 1024m$^2$, and batch size of 512.}
    \label{tab:results_2}
\end{table}

 \subsection{Feature Representations Evaluation}
 
 \alex{In this section, we present qualitative results of CroCo.}
 
 \paragraph{Heatmap.}
 
 We compute similarity metrics and represent them in a heat map to understand if
 a positive pair of RGB and DEM image learns a comparable feature embedding. If
 the base encoders learn discriminative feature descriptors, then the positive
 pair will give the only high score in the similarity matrix. This means all the
 other patches are far apart in the latent space. If the similarity score for
 the positive pair either doesn’t give the highest score or doesn't stand out
 from the rest, then the descriptor is not discriminative.

\begin{figure} \centering
    \includegraphics[width=\linewidth]{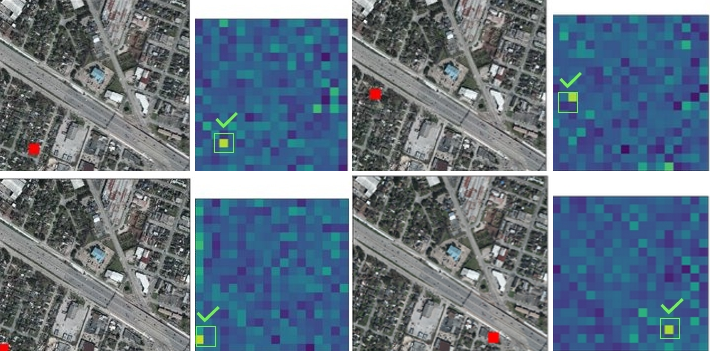}
    \includegraphics[width=\linewidth]{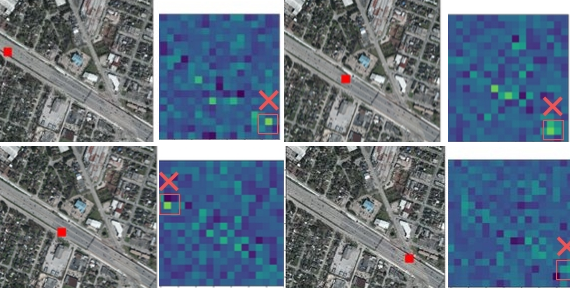} \caption{Heat map
    generated from the similarity matrix between the target patch DEM and all
    RGB patches.} \label{fig:heat_map} \end{figure}

In Fig~\ref{fig:heat_map}, we randomly plot some heat maps that visualize the
similarity matrix of the target DEM patch to all RGB patches. The red box on the
aerial image on the left denotes our target patch. The figure on the left is the
similarity score between the query DEM descriptor against all the RGB
descriptors. The value on the heat map represents the similarity score at the
corresponding position. For example, the pixel at the upper left corner is the
similarity score between the target patch’s DEM descriptor and the upper left
most RGB patch’s descriptor. If the highest score on the heat map is indeed from
the positive pair of the target, then the base encoder manages to generate a
discriminative feature vector for that target. The query DEM patch is thus
correctly localized.  

We give eight examples of localization attempts, the upper four are successful
cases and the lower four are failures. In the successful cases, the target gives
the highest score that strikes a clear difference from the rest patches. There
are also some pixels with relatively higher scores corresponding to patches with
similar structure and features as the target patch. Despite that, our method
successfully retrieves the positive pair. As for our failed attempts, the
mistakes mostly occur on the large traffic lane and freeway. From the heat map,
we can see a strip of the bright zone that stretches along the direction of the
lane. This reflects localizing in a monotonic area with less dynamic topography
is difficult with our approach. As we attempt to map bird-eye view DEM to the
RGB imagery, this makes it exceptionally challenging for flat areas. Considering
the ground view, extra information from the street side might help with
overcoming the shortage. 

\paragraph{Localization outcome.}

\begin{figure}[t] \centering \includegraphics[width=\linewidth]{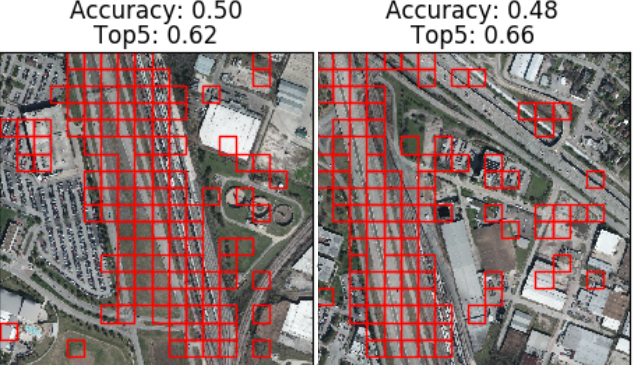}
\caption{Evaluation on validation dataset.} \label{fig:val_eval} \end{figure}

\begin{figure}[t] \centering \includegraphics[width=\linewidth]{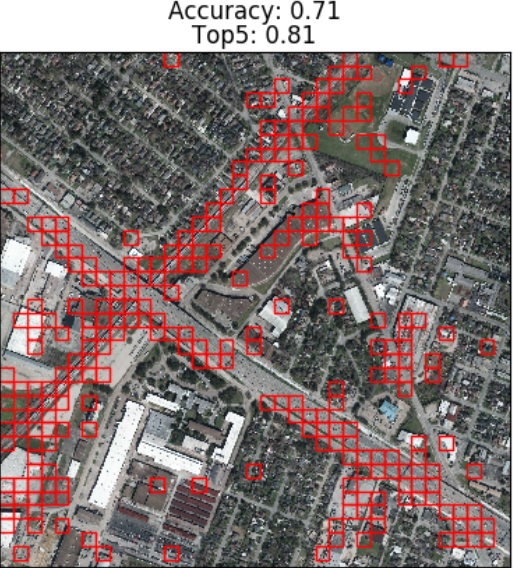}
\caption{Evaluation on test dataset.} \label{fig:test_eval} \end{figure}

\begin{figure}[t] \centering
\includegraphics[width=\linewidth]{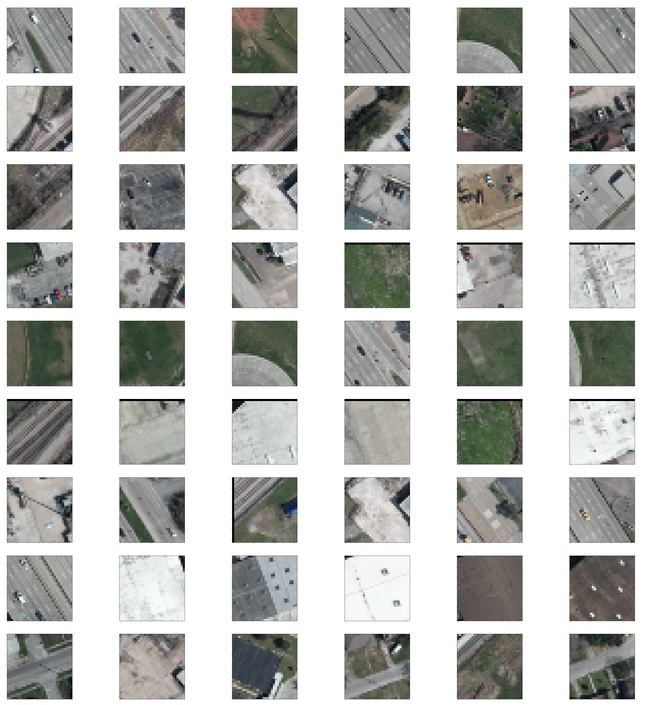}
\caption{Target RGB patches (left column) and the wrong Top-5 retrievals (following patches).} \label{fig:wrong_top5}
\end{figure}

The percentage of corrected localized patches, as already explained, is denoted as the Top-1 score. Fig.~\ref{fig:val_eval} and
Fig.~\ref{fig:test_eval} visualize the distribution of correct and incorrect
retrievals over the map. The patches labeled with red boxes are those with the
wrong Top-5 retrievals. More errors happen in flat areas such as highways,
traffic lanes, rooftops, or grass areas. The phenomenon occurs in both
validation and test sets. The result indicates DEM data is insufficient to give
a solid localization in the area without much terrain changes. When the
landscape is more diverse with complex formation, the encoder can extract
invidious features and performs better localization.  

In Fig.~\ref{fig:wrong_top5}, we randomly select some patches that fail to
retrieve the correct RGB patch in the Top-5 highest scores. The first column is
the target RGB and the following five patches are the retrieved patches. We
observe that the retrieved images mostly show a very similar structure and
content with the target. For example, in the first two instances, all retrieved
shape indicates a similar alignment with the target patch. Also for the instance
5 and 6, retrieval patches are plan areas without extra features to refer to.
Our query data are DEM patches with information about the surface or terrain
elevation, which lacks color information as in RGB images. Therefore, some
retrieval errors that might have been easily ruled out with visual clues become
difficult to avoid with the use of only a digital terrain model. 

\subsection{Discussion and perspectives}
\label{sec:discuss}

\alex{
\paragraph*{Training data.}
As aforementioned, the size of the training dataset is of importance.
We followed the same procedure to generate the training set as we did for generating the evaluation set.
We use the NT-Xent loss~\citep{Chen2020} originating from self-supervised literature.
It appears that the size of the training set along with the possible data augmentation known as critical for self-supervised learning,  shall also be investigated.
It is a promising perspective to deeply investigate how to enhance our training set in order to boost the performances.

\paragraph*{Using time and context in the analysis.}
We tackle the relocalization problem as a static one patch problem.
In practice, acquisitions are often sequential and using the location of the previous images can help enhancing the precision of the localization and discarding the false positives.
Moreover, it would allow us to create trajectories.
}

\section{Conclusions} \label{sec:conclusions}

In this research, we proposed a contrastive learning-based method to solve the
visual place recognition problem. We adapt the SimCLR framework as feature
encoders trained on the contrastive loss. The augmentation operator in the
original framework is replaced with our data preprocessing pipeline. Our
contributions in this work are 1) Learning feature descriptors directly from
contrastive learning architecture for image matching and retrieval tasks. Our
method learns the feature descriptors for the RGB branch and the DEM branch
simultaneously and could be used for coarse localization right after training.
2) Localize the rasterized point cloud data on the aerial image. Ground-based
prior map generation and maintenance are resource-intensive. Oppositely, remote
sensing-based prior maps have dense coverage, high-resolution, and are
well-maintained, geo-referenced. With the use of pre-aligned aerial images and
rasterized point cloud, training samples can be created simply without
additional annotation work. 

We conduct experiments on different sampling strategies and hyper-parameters.
The proposed framework is able to learn discriminative features from RGB and DEM
pairs.
Despite the differences, our
contrastive learning framework has demonstrated the feasibility of mapping
positive data pairs into close feature embeddings for retrieval tasks.
In general, the trained encoder learns to map the
features correctly, and the similarity score at the target patch is notably
higher than all other patches.

Our experimental result indicates that (with other factors fixed) higher image
resolution gives better accuracy. However, one thing we noticed during our
experiments is that with limited computation resources, the lower resolution
allows a larger dataset and batch size.
To sum up, resolution and computation power is compensated when
applying this method for image retrieval tasks. The best performance does not
necessarily come from higher image resolution with the constrain of computation
power.

In correspondence to the conclusion for SimCLR, contrastive learning benefits
from a larger batch size. This is because the contrastive loss is computed
against all the negative pairs within the same batch, and a larger batch size
creates more negative pairs. In our experiments, we notice that a larger batch
size gives better Top-1 and Top-5 scores. As for the sampling stride selection,
the opposite tendency is observed in 50-cm resolution and 100-cm resolution
scenarios – smaller stride gives better accuracy for the former, but worse for
the latter. We think it takes some further analysis to understand how the
sampling stride impacts the result.

The strengths of the proposed methods are as follows: first, it is a learning-based feature,
meaning no need for hand-crafting or domain-specific knowledge; second, it deals
with an arbitrary input data type. The framework learns to generate identical
feature vectors from the positive pairs. In our case, we define a pair of RGB
and DEM imagery as our data pair. However, the input data is flexible and free
of choice as long as reasonable data pair definitions are given. The weakness of
the proposed method is the poor performance in localizing patches to an area
with not much feature or terrain change. This is restricted by the limited
information a DEM can provide. A potential solution to mitigate this weakness is
to include extra data that give more context information such as the texture and
color (the latter can be extracted from multispectral LiDAR sensors).

In this work, we successfully implement a contrastive learning framework to
solve the localization problem using aerial RGB imagery and a rasterized point
cloud. We explore different data pre-processing methodologies and also fine-tuning of 
the hyper-parameters. In our best scenario, we achieved 0.71 for the Top-1 score
and 0.81 for the Top-5 score on the test dataset. With the heat map to visualize
the similarity score between the target patch and the entire dataset, we see
that the feature descriptors are discriminative and can efficiently retrieve the
target patch. The proposed framework can take an arbitrary base encoder. Some
future directions would be to change the backbone architecture, like a different
version of ResNet. In our work, we tried only one strategy to concatenate the
digital terrain models. Further analysis can be done to compare with other
choices of the input data. The input data is not restricted to 2D imagery, as
contrastive learning simply learns the embedding based on user-defined positive
data pairs. As an extension of this research work, we are now working further on
3D point cloud data as input and corresponding 3D CNN as a base encoder.

\section*{Acknowledgements}
The authors would like to thank the National Center for Airborne Laser Mapping and the Hyperspectral Image Analysis Laboratory at the University of Houston for acquiring and providing the data used in this study, and the IEEE GRSS Image Analysis and Data Fusion Technical Committee.

\renewcommand{\bibsection}{\section*{References}}
\bibliography{IRISA-ISPRS_EV}

\end{document}